# Examplers based image fusion features for face recognition

Alex Pappachen James*[1] and Sima Dimitrijev[2]

*[1] Asst. Professor and Group Lead, Machine Intelligence Group, Indian Institute of Information Technology and Management-Kerala, India. www.mirgroup.co.cc, apj@ieee.org

[2] Professor and Deputy Director,Queensland Micro- and Nanotechnology Center, Griffith University, Australia, www.gu.edu.au/qmnc s.dimitrijev@griffith.edu.au

## Abstract

Examplers of a face are formed from multiple gallery images of a person and are used in the process of classification of a test image. We incorporate such examplers in forming a biologically inspired local binary decisions on similarity based face recognition method. As opposed to single model approaches such as face averages the exampler based approach results in higher recognition accuracies and stability. Using multiple training samples per person, the method shows the following recognition accuracies: 99.0% on AR, 99.5% on FERET, 99.5% on ORL, 99.3% on EYALE, 100.0% on YALE and 100.0% on CALTECH face databases. In addition to face recognition, the method also detects the natural variability in the face images which can find application in automatic tagging of face images.

## 1. Introduction

Memory in humans enables the functions of intelligence and learning. In the case of face processing [4], memory is required during the stage of comparison of an identity information from a test face to that of the stored identity information of any gallery face, in fact, this is true for any knowledge based learning. Usually, the identity information from a test image is compared with all the stored representation of faces in the gallery set. The best match of these comparisons results in the identification of a face. It can be observed that in such a situation as long as the images in the gallery set are not removed the class of a test image can be identified with some accuracy using any predefined face recognition algorithm. The two basic blocks of a machine based face recognition algorithm are feature extraction and classification. Feature extraction process aims at the extraction of maximum information, while classification tries to find best discrimination between the comparing class of faces.

The main contributing factors that make the recognition of faces difficult are the limitation in training data per person and the natural variability between the images in the gallery set and the test set. Various types of natural variabilities that make the comparison between a test image with that of the gallery images are: (1) localization mismatches, (2) illumination variation, (3) expression of faces, (4) internal or external occlusions, (5) variations in pose of faces and (6) ageing of faces with noise [2, 19, 13, 14, 7, 18]. To have a face recognition algorithm that is invariant to all such variabilities is a very difficult task. There are many possibilities and ways of handling such variabilities: (1) one can think of dealing with each variability individually and construct a feature vector that is invariant to all the known variabilities, (2) form a robust feature extraction method that is invariant to most of the variability and then improve the system performance by other known compensation techniques, and (3) use multiple images to form the gallery images of a person such that all the possible variabilities are accounted for.

In humans it is possible that all of these techniques are employed in a way or other and it becomes logical to try and implement the combination of techniques in the implementation of a face recognition algorithm in machines. In an attempt to emulate this, we developed a biologically inspired face recognition method based on local binary decisions and spatial intensity change filtering. In past we successfully implemented these methods for single gallery image per person problem [5]. The high recognition performance against such difficult task enables us to explore further the effect of using multiple gallery images per person to form the gallery set. The increased number of gallery images per person provides more information and hence one can expect higher recognition accuracies than when using a single gallery image per person. However, there are different ways in which the gallery data can be used for comparison: (1) by forming a single feature model (e.g. by taking average) from the set of gallery images of a person [6], and (2) use selected gallery images individually as examplers for comparison. The first method is a simple average model and the second is a exampler model.

In this paper, we show the recognition performance of a face recognition method that uses local binary decisions on similarity of features for recognition of faces when there are ordered and unordered multiple number of gallery images per person in the gallery set. The average model and exampler models are used as a training models for comparison of test image with the gallery images. Further, we compare the recognition performance of the designed algorithm against other known best performing algorithms. Also, we analyze the effect of multiple gallery images on localization error compensation and dimensionality reduction.

2. Face recognition method

The local binary decisions on similarity method consists of a feature extraction and a local binary decisions based classifier [5]. In the proposed algorithm, the feature extraction stage consists of texture based spatial filters followed by spatial change detection filters and the classifier applies local binary decisions on an average similarity measure. The use of multiple images can be incorporated by either using a single average model for multiple images or a direct exampler model.

2.1. Feature extraction

Any raw intensity image of size N × M pixels can be denoted as $I(i, j)$, where $(i, j)$ represents the location of pixel. Further, we denote the gallery images as $I_g^{(d,k)}$, where $d = \{1 \ldots D\}$ is the index for a person, and $k = \{1 \ldots K\}$ is the index of the sample images of the same person. Similarly, a test image is represented as $I_t^{(d,*)}$, where $*$ denotes any random index of sample.

Both gallery and test images require to go through the feature extraction stage. As part of feature extraction, we use texture based spatial filtering as the preprocessing operation that can maximize the availability of identity information in a face. The output from the texture based spatial filters are then subjected to spatial intensity change detection operation using local standard deviation filter. This completes the feature extraction stage of the proposed algorithm.

Texture based spatial filtering is achieved by a linear convolution between a specified filter window coefficients w and raw image $I$. This operation is expressed as:

$$Y^{(p,*)}(i, j) = \sum_{s=-a0}^{} \sum_{t=-b0}^{} w^{(p)}(s, t) I(s+i, t+j) \qquad (1)$$

where, $a0 = (n0 - 1)/2$, $b0 = (m0 - 1)/2$, $p = \{1, \ldots, P\}$, P is the maximum number of the filters applied and w has a size of m0 × n0 pixels. This is followed by spatial change detection using a local standard deviation filter across a window of n × m pixels summarised as:

$$\sigma^{(p,*)}(i, j) = \sqrt{\frac{\sum_{s=-a}^{b} \sum_{t=-b}^{b} [Y^{(p,*)}(i+s, j+t) - \overline{Y^{(p,*)}(i,j)}]^2}{mn}} \qquad (2)$$

where $a = (m - 1)/2$, $b = (n - 1)/2$ and $\overline{Y^{(p,*)}(i, j)}$ is the local mean.

Normalization is performed on Eq. (2) to form the final feature vector:

$$x^{(p,*)}(i, j) = \sigma^{(p,*)}(i, j) / \overline{\sigma^{(p,*)}(i, j)} \qquad (3)$$

where, the local mean $\overline{\sigma^{(p,*)}(i, j)}$ is calculated on a window of features of size k × l pixels.

We use six texture based spatial filters in the proposed method. These filters and its corresponding outputs are shown in Fig. 1 (a)-(f), while its corresponding normalized spatial change is shown in Fig. 1 (g)-(l).

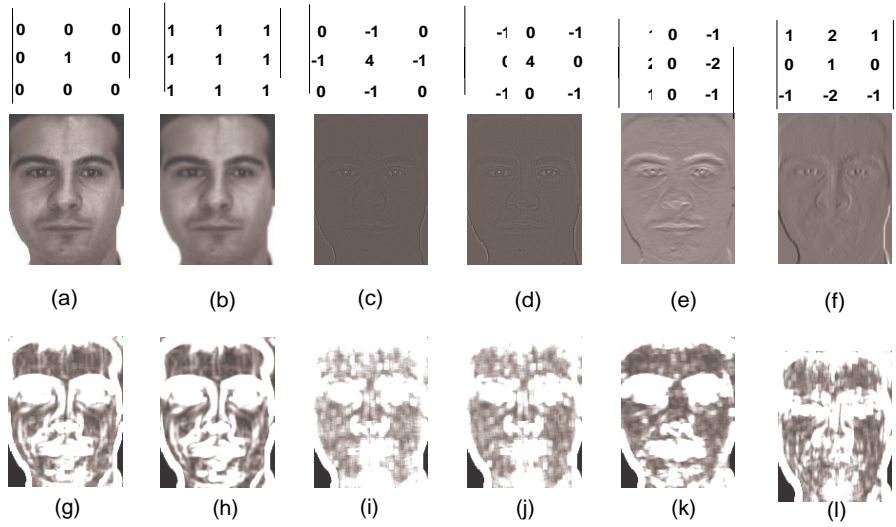

Figure 1: Illustration of texture based spatial filtering applied to a face image in AR database. The weights of the six texture based spatial filters and its corresponding outputs are shown in images labeled (a) to (f). The spatial change detection is applied on each of these outputs and is shown in images labeled (g) to (l).

2.2. Classification

The classification process involves the comparison between the feature vectors of a test image with that of gallery images. One of the simplest way to form such comparison is by taking absolute difference between the features, and we improve this difference method by normalization and we call the resulting vector as similarity measure vector. This feature comparison operation is summarized as:

$$\delta^{(p,d,k)}(i, j) = \frac{|x_g^{(p,d,k)}(i, j) - x_t^{(p,*)}(i, j)|}{\min(x_g^{(p,d,k)}(i, j), x_t^{*}(i, j))} \quad (4)$$

A comparison of a test feature vector with a gallery image having a index d will result in p similarity measure vectors due to the p texture based spatial features employed during the feature extraction process. To simplify the calculation and to reinforce the identity information the p similarity measure vectors are averaged to form a single average similarity measure vector given by:

$$\hat{\delta}^{(d,k)}(i, j) = \frac{1}{P} \sum_{p=1}^{P} \delta^{(p,d,k)}(i, j) \quad (5)$$

Application of local binary decisions on $\hat{\delta}^{(d,k)}$ using a global threshold $\theta$ results in a binary decisions vector $B^{(d,k)}$ represented as:

$$B^{(d,k)}(i,j) = \begin{cases} 1 & \hat{\delta}(i,j) < \theta \\ 0 & \hat{\delta}(i,j) \geq \theta \end{cases} \quad (6)$$

This binary decisions vector is used form a global similarity score $S_g^{(d,k)}$ given by:

$$S_g^{(d,k)} = \sum_{i=1}\sum_{j=1} B^{(d,k)}(i,j) \quad (7)$$

The comparison of a test image with d × k gallery images results in d × k similarity scores and the maximum value among all such scores will represent the best match of the test image in the gallery:

$$dk^* = \arg\max_{d,k} S_g^{(d,k)} \quad (8)$$

### 2.3. Compensation of localization errors

Another, important practical aspect in face recognition is the localization error that occur due to any improper detection and alignments. In most practical cases, a perfect alignment is difficult to achieve, so it is logical to add a scheme to compensate for such errors. In practice, localization of faces are done with respect to eye coordinates of the face image in majority of face detection schemes. This knowledge is used to form a simple compensation method by perturbating location of eye coordinates to generate scaled, shifted and rotated versions of an image. These perturbations can be applied to the gallery or the test image and compensates for variations in shifts, scale and rotation. Any l number of perturbations on an image will result in l similarity measure, and the one with the maximum value is chosen as the representative score for that comparison. In this way small variations in localization errors that can reduce the recognition performance is compensated.

### 3. Experimental Results

#### 3.1. Equal number of training samples per person

One particular situation in multiple training samples per person problem occurs when there are equal number of training samples per person with same conditions of natural variability for all the persons in the gallery. This method of setting up the gallery is perhaps the easiest way to achieve high recognition performance due to the equal probability of match on the identity information of a test image with that in the gallery images. Further, within this scheme, the gallery samples can be setup in two ways: (1) use all the training samples per person directly for comparison (exampler method), and (2) form a single model gallery image for a person from all the training samples for a person (average method).

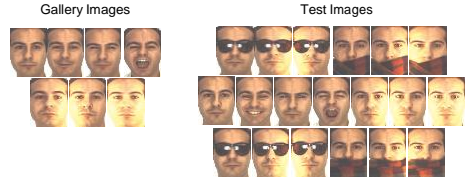

Figure 2: Illustration on the organization of AR database [11, 12] used for testing. The galley images consists of first 7 images taken on session one containing expressions and illuminations. The test images consists of 6 images taken on session one and 13 images taken on session two and contains occlusions, expressions and illumination.

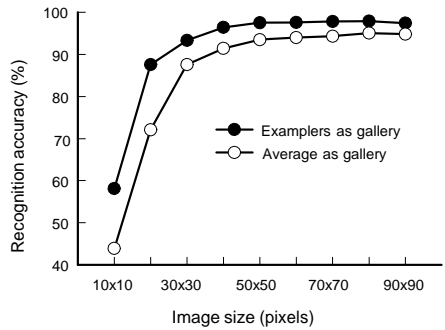

Figure 3: A graphical illustration of the recognition performance of the proposed method with variation in dimensionality of feature vectors. This shows the specific situation when the number of training samples used for creating the gallery is same for all the persons. AR database is employed for these simulation. Further this also shows a comparison of single model approach such as average feature model against multiple feature models such as exampler features.

For the experimental analysis, on all the databases following are the values of parameters used in the algorithm for the value of image/feature vector size kept at a range from 10 × 10 pixels to 90 × 90 pixels: (1) standard deviation filter size is 3 × 3 pixels, (2) local mean normalization window size for forming normalized features is kept at 30 × 30 pixels, (4) the value of global threshold is 0.25 and perturbations of ±5 pixels are applied in horizontal, vertical and diagonal directions.

We use AR database [11, 12] with images of 100 persons for the simulations reported in this paper. For each person there are 26 different images representing 13 conditions over two sessions. Seven images of each person are selected as training samples to form the gallery (see Fig. 2) and the remaining 19 images of each person is used for testing the recognition performance of the algorithm.

From Fig. 3 it can be seen that exampler method performs better than average method for all the shown variation in the dimensionality. However, exampler method require k times more amount of memory and comparisons as opposed to average method and hence results in larger computational time as shown in Fig. 4. Further, it can be seen from Fig. 3 that exampler method shows robust recognition performance than average method with respect to variations in dimensionality of feature vector.

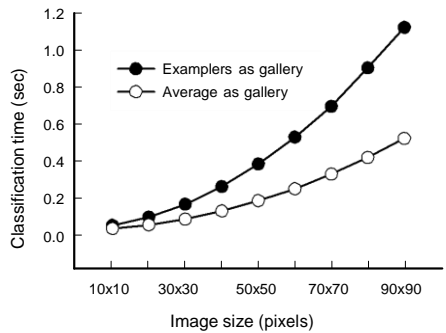

Figure 4: Graphical illustration showing the time it takes for the classifier for a comparison of a test with all images in the gallery at different feature vector dimensions.

Table 1 shows the recognition performance and overall robustness of proposed method using color images and perturbations. For AR [11, 12], ORL , FERET [16, 17], EYALE [8, 3], YALE [1] and CALTECH databases the reported results are for the feature size of 60 × 60 pixels, 40 × 40 pixels, 60 × 60 pixels, 80 × 80 pixels, 60 × 60 pixels and 80 × 80 pixels. ORL database which contains pose variation of 40 persons is also tested to benchmark the performance, and clearly show high recognition performance. FERET database with images of 200 persons are used and each person has 3 photos representing different natural variability. EYALE database has 64 photos with different illumination variation on each of the 10 persons in the database. This database base is often used to benchmark the face recognition performance against serious illumination variations. YALE database has photos of 15 persons under 11 different natural variability. CALTECH face database has photos of 28 persons under random background and in natural lighting conditions. Clearly, from Table 1 the proposed method show high recognition performance in all the tested databases, which confirms the overall robustness of the method.

3.2. Unequal number of training samples per person

Yet another problem in face recognition with multiple training samples per person is formed when number of training samples per person is different. However, this problem is closely related to how learning occurs in primates where the availability of information for training a face is different from another. Furthermore, for majority of biometric application such scenarios are mostly expected to occur.

Figure 5 shows the effect of monotonic increase in training samples in creating a gallery set for the all the persons. It can be clearly seen that an increase in available information during training increases the recognition accuracy considerably. Further, this also mean that the increased used of memory for storing any distinct identity information results in better recognition performance and stability. This is substantiated from the simulation results shown in Fig 6. These results show that using exampler method, involving multiple representation of a face provides greater stability and recognition performance as opposed to a single model approach.

Table 1: Recognition performance across different databases on the proposed method when the number of training samples in the gallery and test are fixed for every person.

| Database | Samples per person | | Recognition Accuracy(%) Perturbation | |
|---|---|---|---|---|
| | Gallery | Test | Yes | No |
| AR | 7 | 19 | 99.0 | 97.9 |
| ORL | 5 | 5 | 99.5 | 97.5 |
| FERET | 2 | 1 | 99.5 | 92.5 |
| EYALE | 7 | 57 | 99.3 | 99.3 |
| YALE | 5 | 6 | 100.0 | 100.0 |
| CALTECH | 5 | 1-15 | 100.0 | 100.0 |

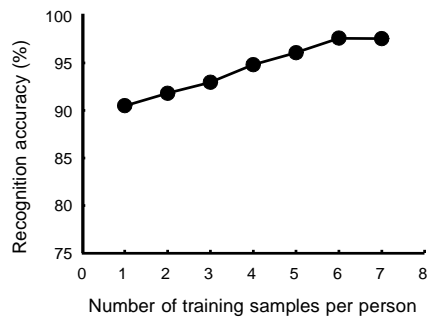

Figure 5: Graphical illustration showing the effect of using multiple training samples. The simulation is done by exampler feature comparisons using gray scale images of size 60×60 pixels in the AR database and the proposed method is used without compensating for localization errors.

3.3. Comparison with other methods

We compare the proposed method using examplers with other best performing and well known algorithms reported in the area of face recognition. This comparison is organised based on the databases used to report the result.

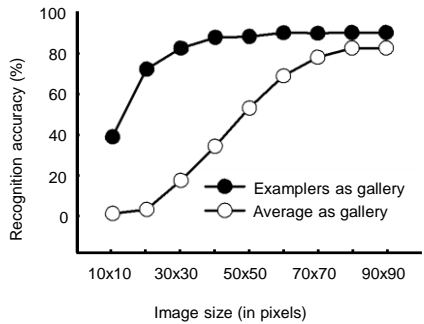

Figure 6: A graphical illustration of the recognition performance of the proposed method with variation in dimensionality of feature vectors when the number of training samples used for creating the gallery are randomly selected and are not same for all the persons. Using AR database this simulation also shows a comparison of average feature model against exampler feature model.

3.3.1. AR Database

The following methods are used for comparison against the proposed proposed algorithm: SIS [10], LDA [1], DCV [9] and PCA [15]. Clearly, from Table 2 it can be seen that proposed algorithm outperforms all other methods. In this simulation 7 images per person is used for forming the gallery and 19 images per person are used for testing. It can be also noted that AR database contains occlusions of faces which is often considered as a difficult task for recognition. Also this simulation shows the robustness of the proposed algorithm against different natural variability under the same pose.

3.3.2. ORL, FERET and YALE Databases

As seen from Table 2, MGFR-2D$^2$PCA [20] and 2D$^2$PCA [21] are used for comparison against presented method. First five images of each person in ORL database is used as training samples, while the remaining five are used for testing. In case of FERET, any 2 images are selected as the training sample, while the remaining image forms the test sample. For each person in the Yale database, we use the first 5 images to form the training sample and the remaining 6 images forms the test samples. It can be seen from the reported results that the presented method has very high recognition accuracy and is robust against different databases. Further, it is worth while to mention that in the case of YALE database 100 percent recognition accuracy is achieved even with single training sample per person. In terms of natural variability these databases include variations in pose, expressions and illumination.

3.3.3. EYALE Database

Results on EYALE database (see Table 2) shows the recognition performance of the presented method against other best performing methods such as 9PL$_{real}$ [8], Cone cast [3] and SIS [10]. In each of these methods, the images in subset

Table 2: The comparison of recognition performance of the proposed method for multiple training samples person face recognition problem with other algorithms.

| | Top rank recognition accuracy (%) | | | | |
|---|---|---|---|---|---|
| Database | Proposed | SIS | LDA | DCV | PCA |
| AR | 99.0 | 98.1 | 56.5 | 57.4 | 44.7 |
| | Proposed | MGFR-2D$^2$PCA | 2D$^2$PCA | | |
| ORL | 99.5 | 100.0 | 92.5 | | |
| FERET | 99.5 | 99.5 | 92.5 | | |
| YALE | 100.0 | 98.9 | 91.1 | | |
| | Proposed | SIS | Cones-cast | 9PL$_{real}$ | |
| EYALE | 99.3 | 99.7 | 100 | 100 | |

1 consisting of 7 images with a zero degree angle of the light-source directions from the optical axis are used as the training samples. Further, it can be noted that the shown results on cone cast and 9PL$_{real}$ are using first four subsets, while for SIS and proposed all the five subsets are used. Clearly, from Table 2 the comparison with well known method like cone cast and 9PL$_{real}$, the presented method shows very similar recognition performance. This also shows the near invariant recognition performance of the proposed method against serious illumination changes.

3.4. Knowledge based recognition

Examplers also provide a knowledge on condition or natural variability. This prior knowledge of the gallery examplers helps with detections of facial expressions, variations in time, aging and pose. In order to demonstrate this, 13 images taken on session 1 in the AR database is used as the exampler gallery images, while the remaining 13 images from session 2 are used as test. The 13 images are grouped based on the knowledge of the conditions as: (1) Neutral, (2) Expression, (3) Illumination, (4) Eye occlusion and (5) Mouth occlusion. The test image when compared with the images in the gallery results in a set of similarity scores. These scores are ranked and the associated natural variability is tagged to the test image.

Table 3 shows the recognition performance of the proposed face recognition algorithm in the detection of natural variability in the image. The detection of occlusion at rank one seems to be most difficult task. This is due the fact that mouth occlusion images contain the least amount of identity information and will result in lower values of similarity score. This is a relative disadvantage of using a global similarity score, however a possible way to improve the recognition performance is by using with local approaches such as by the calculation of region-wise similarity scores and weights. The ability of the proposed method to detect natural variability and recognise faces with high accuracies makes it useful in the various applications of automatic tagging of face images.

Table 3: Recognition accuracy of the proposed algorithm in the detection of natural variability in the face images.

| Rank | Recognition accuracy (%) | | | | |
| --- | --- | --- | --- | --- | --- |
| | Neutral | Expression | Illumination | Eye occlusion | Mouth occlusion |
| 1 | 83.0 | 87.4 | 96.0 | 99.0 | 73.0 |
| 2 | 98.0 | 100.0 | 99.0 | 100.0 | 89.7 |
| 3 | 99.0 | 100.0 | 99.7 | 100.0 | 96.7 |
| 4 | 100.0 | 100.0 | 100.0 | 100.0 | 100.0 |

4. Conclusion

In summary, we presented an exampler based local binary decisions on similarity algorithm that we successfully applied to multiple training samples per person face recognition problem. We showed the relative advantage of using examplers in comparison with single model approach such as by averaging. In the case of exampler based approach, the increased usage of memory enables the use of more identity information as opposed to single model approach. Although single model approach is computationally less expensive, the use of examplers enables a stable performance even with reduced feature dimensionality. The presented method outperforms other major algorithms in overall robustness across various natural variabilities. This is attributed to the use of texture based spatial change features and the use of local binary decisions classifier. Further, the use of multiple training samples helps in compensation of natural variability and increases the probability for a true match. In addition, a useful aspect of this method is its ability to detect natural variability which sets this method apart from its counterparts. Finally, from the results it is evident that any increase in the number of examplers will make the recognition performance higher and more stable across variations in natural variability and feature dimensionality.